\documentclass[sigconf]{acmart}

\AtBeginDocument{%
  \providecommand\BibTeX{{%
    \normalfont B\kern-0.5em{\scshape i\kern-0.25em b}\kern-0.8em\TeX}}}

\usepackage{algorithm}
\usepackage{algorithmic}
\usepackage{multirow}

\renewcommand\footnotetextcopyrightpermission[1]{}
\begin{document}
\acmConference[Arxiv Version]{}{August 2023}

\title{A Symbolic Character-Aware Model for Solving Geometry Problems}

\author{Maizhen Ning}
\orcid{0000-0002-8842-4187}
\affiliation{%
  \institution{Xi’an Jiaotong-Liverpool University}
  \city{Suzhou}
  \country{China}
}
\email{maizhen.ning16@student.xjtlu.edu.cn}

\author{Qiu-Feng Wang}
\orcid{0000-0002-0918-4606}
\authornote{Corresponding author.}
\affiliation{%
  \institution{Xi’an Jiaotong-Liverpool University}
  \city{Suzhou}
  \country{China}}
\email{qiufeng.wang@xjtlu.edu.cn}

\author{Kaizhu Huang}
\orcid{0000-0002-3034-9639}
\affiliation{%
  \institution{Duke Kunshan University}
  \city{Suzhou}
  \country{China}
}
\email{kaizhu.huang@dukekunshan.edu.cn}

\author{Xiaowei Huang}
\orcid{0000-0001-6267-0366}
\affiliation{%
  \institution{University of Liverpool}
  \city{Liverpool}
  \country{UK}
  }
\email{xiaowei.huang@liverpool.ac.uk}

\begin{abstract}
    AI has made significant progress in solving math problems, but geometry problems remain challenging due to their reliance on both text and diagrams. In the text description, symbolic characters such as "$\triangle$ABC" often serve as a bridge to connect the corresponding diagram. However, by simply tokenizing symbolic characters into individual letters (e.g., 'A', 'B' and 'C'), existing works fail to study them explicitly and thus lose the semantic relationship with the diagram. In this paper, we develop a symbolic character-aware model to fully explore the role of these characters in both text and diagram understanding and optimize the model under a multi-modal reasoning framework. In the text encoder, we propose merging individual symbolic characters to form one semantic unit along with geometric information from the corresponding diagram. For the diagram encoder, we pre-train it under a multi-label classification framework with the symbolic characters as labels. In addition, we enhance the geometry diagram understanding ability via a self-supervised learning method under the masked image modeling auxiliary task. By integrating the proposed model into a general encoder-decoder pipeline for solving geometry problems, we demonstrate its superiority on two benchmark datasets, including GeoQA and Geometry3K, with extensive experiments. Specifically, on GeoQA, the question-solving accuracy is increased from 60.0\% to 64.1\%, achieving a new state-of-the-art accuracy; on Geometry3K, we reduce the question average solving steps from 6.9 down to 6.0 with marginally higher solving accuracy. The code is available at \href{https://github.com/ning-mz/SCA-GPS}{https://github.com/ning-mz/SCA-GPS}.
\end{abstract}


\maketitle

\begin{figure}[t]
  \centering
  \includegraphics[width=\linewidth]{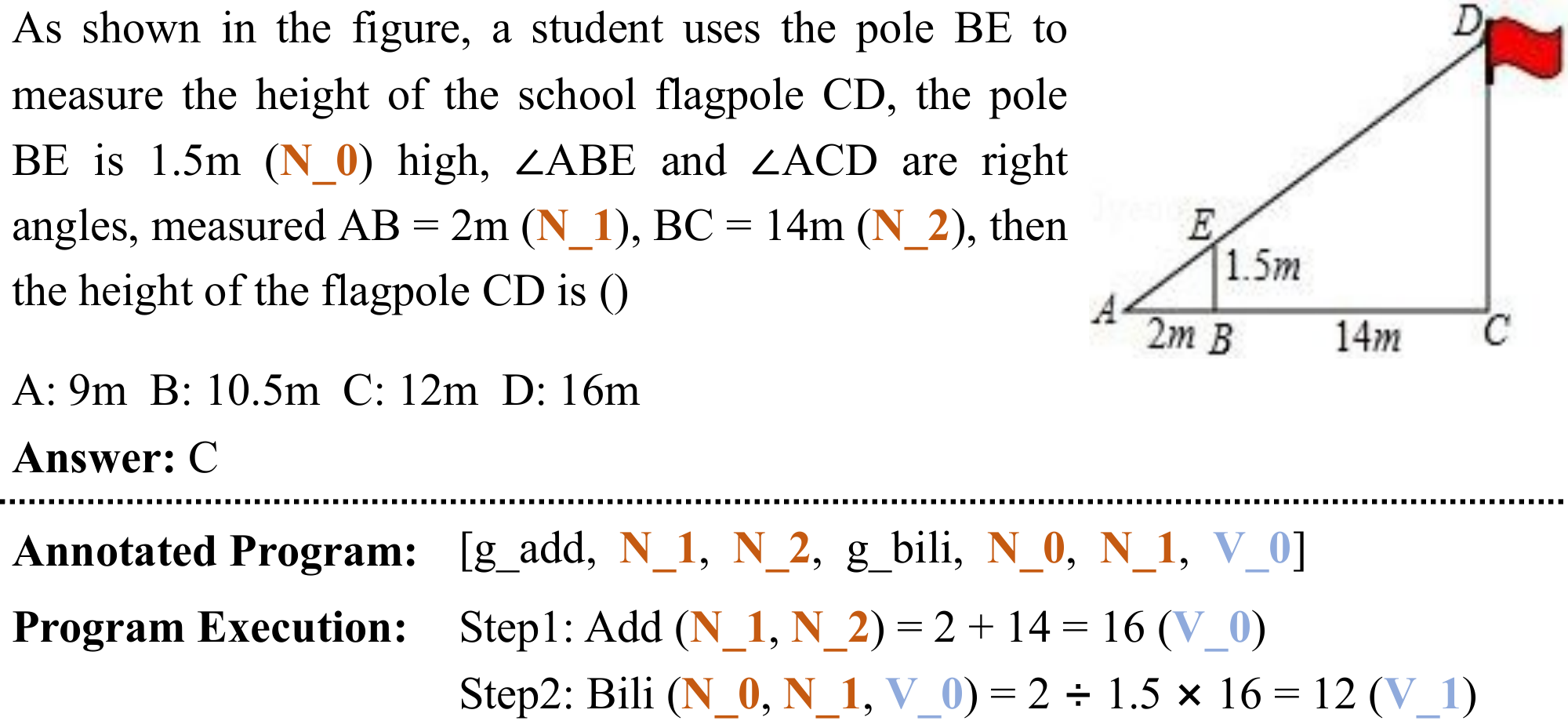}
  \caption{A typical geometry problem with the annotated program in the GeoQA dataset \cite{GeoQA}. The top part shows a geometry problem with the answer, and the bottom part gives the annotated program (i.e., the ground-truth sequence) and how the program is executed to obtain the final answer.}
  \label{fig:example}
\end{figure}

\section{Introduction}
Solving math problems with AI techniques is meaningful in automatical system reasoning and math education, which has attracted much attention and made great progress with the flourishing development of deep learning techniques~\cite{bajaj2018smart,goalTreeDec,graph2tree,acl-ours}. However, most existing works  focus on solving math word problems (MWP) that are mainly arithmetic and algebraic based on text description only~\cite{introMWP1,shen2021generate,seekingPattern,mwpAsIE}. In this paper, we aim to solve geometry problems (GP), an even more challenging yet valuable task~\cite{introEdu}.

Solving GPs essentially requires joint reasoning on multimodal data~\cite{GeoQA}. As illustrated in Figure \ref{fig:example}, a GP typically consists of two modalities, i.e. text description and a diagram image. In contrast to the MWP which only has textual input, the GP contains a geometry diagram, which is complementary to texts. Therefore, on top of the text encoder, the GP solver is usually equipped with a diagram encoder to capture geometric information. 
Then a multi-modal joint reasoning module is necessary to align both encoded textual and visual diagram features, where the symbolic characters often serve as a crucial linkage between problem text and diagram, e.g., "AB", "BC", "CD" and "BE" in Figure \ref{fig:example}. GPs turn out more difficult to be solved, since the final answer (e.g., $12m$ in Figure~\ref{fig:example}) provides rather limited information to learn and requires complex theorem knowledge for reduction. To this end, researchers usually design a sequence of handcrafted math calculation program codes (e.g. the annotated program in Figure \ref{fig:example}), which are executed to output the final answer of the problem. Meanwhile, it is important to learn some geometric theorems in diagram understanding, e.g., the Proportion theorem should be explored to calculate the length "CD" in Figure~\ref{fig:example}. 

Prior works on solving GPs highly depend on handcrafted rules to parse the problem, which may not be accurate enough and only work on small-scale data~\cite{seo2014diagram,sachan2017learning}. Moreover, these rule-based methods are hard to be applied for diagrams with variable styles~\cite{DiscourseinMultimedia}. 
With the development of deep learning, researchers start to explore multimodal analysis in solving GPs. 
\cite{GeoQA} introduced a large dataset and an encoder-decoder framework, with separate encoders to extract text description and diagram features, followed by a decoder network predicting program sequences and obtaining final answers through sequence execution.
To better leverage the textual features, \cite{GeoQA+} proposed a dual text encoder model by combining Bi-LSTM and RoBERTa. However, current works simply exploit general NLP models and tackle the text description as a common natural language. This wisdom may not be able to handle diverse geometry symbolic descriptions (e.g. "$\triangle$ACD"), which are highly correlated to the counterpart diagram but share different geometric meanings in various GPs. In addition, these symbolic characters should be jointly represented as a sole variable, while most present text encoders, unfortunately, tokenize them into individual letters (e.g., ‘A’, 'C' and 'D'), losing the semantic relationship with the diagram.   

To tackle the above-mentioned issues in solving GPs, we propose a novel symbolic character-aware model under a general encoder-decoder framework. 
Specifically, we first align counterpart diagram features with individual symbolic character embedding features in the text encoder to extract visual information about symbolic characters. 
Then, we propose to merge these characters to form one holistic semantic unit, aiming to enhance the comprehensive understanding of the geometry description.
Meanwhile, leveraging the Segment Anything Model (SAM) \cite{SAM}, we detect symbolic characters in the diagram and then design a multi-label classification framework to optimize the diagram encoder by predicting the existing symbolic characters.
This task helps the diagram encoder extract more related features to the corresponding text description. 
In addition, we enhance the geometry diagram understanding ability via a self-supervised learning method with the masked image modeling auxiliary task~\cite{vitmae}. 
Extensive experiments demonstrate the effectiveness of our proposed model on two benchmark datasets of GeoQA and Geometry3K.

In summary, the contributions of this paper include:
\begin{itemize}
    \item We propose a novel symbolic character-aware model to fully explore the role of these characters in both text and diagram encoder modules. 
    \item We merge individual symbolic characters to form one holistic semantic unit by aligning counterpart diagram features in the text encoder.
    \item We propose to optimize the diagram encoder under a weakly supervised multi-label classification framework by symbolic characters detection with Segment Anything Model (SAM).
    \item We pre-train the diagram encoder module by a self-supervised learning method with a masked image modeling task. 
    \item We conduct extensive experiments and demonstrate the superiority of our method  on two benchmark datasets of GeoQA and Geometry3K.
\end{itemize}

\begin{figure*}
     \centering
     \includegraphics[scale=0.45]{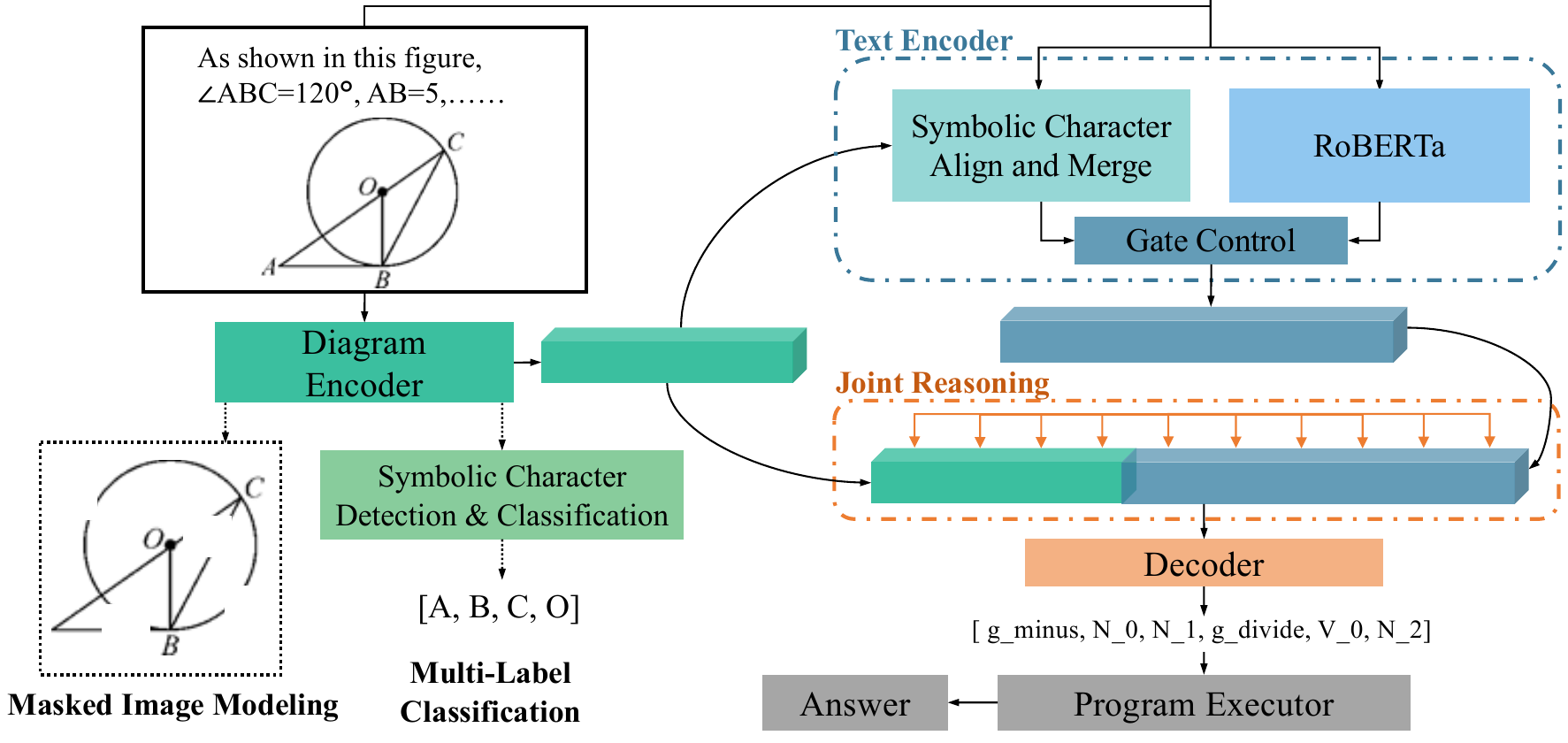}
     \caption{Overview of our SCA-GPS model. Each number in the text is replaced by an independent variable such as $N_0, N_1$, then the text is input to a Text Encoder to obtain text features. In addition, a Diagram Encoder extracts features from the diagram image, which will be fused with text features into the Joint Reasoning block to obtain the final multi-modal features. Next, the final features are input to a Decoder to obtain the program code for the Program Executor, which outputs the final answer. Dash lines mean two auxiliary tasks in the pre-training of the diagram encoder.}
     \label{fig:overview}
\end{figure*}

\section{Related Work}
\subsection{Solving Geometry Problems}
 Solving geometry problems has attracted much attention \cite{rw1960,rw1996} recently. The first automated solver GeoS dealt with SAT-style geometry problems by using NLP and OCR techniques to jointly process text and diagrams into logic forms~\cite{seo2014diagram,introEdu}. 
 Though GeoS attained good performance in GPs, it highly relied on manually designed rules which are hard to cope with various GPs.  As one extension of GeoS, \cite{sachan2017learning} replaced the previous handcraft constraints by engaging horn-clause rules to form the geometry axiomatic knowledge into the solver.
 Recently, Inter-GPS \cite{Inter-GPS} has parsed the problems into formal languages, then applied symbolic reasoning on a series of manually defined complex mathematical axioms. Furthermore, \cite{PGDP} proposed PGDP to separately parse a diagram through non-geometric primitives detection and geometric primitives segmentation, then adopted a GNN \cite{gnn} to model the relationship among primitives. However, both Inter-GPS and PGDP are still rule-based. Their focus is put on parsing geometry problems into formal languages, which unfortunately require complex annotation for training.  Meanwhile, they are hard to be applied in various real GP scenarios, since the parsing module was designed for problems with a single-style diagram and math terminology-based text description (e.g. problems in SAT). For example, the red flag in Figure \ref{fig:example} has no geometry information to be parsed. 
 
 To discover the potential of using deep neural networks in solving GPs, \cite{GeoQA} proposed NGS and applied joint multimodal reasoning on textual and visual features under an encoder-decoder framework. However, the LSTM-based text encoder in NGS is not good at encoding long text descriptions. To tackle this limitation, \cite{GeoQA+} designed a dual text encoder model named DPE-NGS by adopting both Bi-LSTM and RoBERTa \cite{roberta}. 
 However, both NGS and DPE-NGS simply tokenized and encoded the texts as a common natural language, ignoring the semantic meaning of symbolic characters that are closely correlated with the counterpart diagram in different GPs. Differently, in this paper, we propose to explicitly model symbolic characters in both text and diagram encoder.
 
\subsection{Symbolic Characters Detection}
Symbolic characters play an important role as a bridge to connect text description and diagram image. On one hand, previous methods~\cite{Inter-GPS, PGDP} parsed the diagram into textual formal languages to connect the text description. 
However, these methods only perform well for high-quality diagram images with consistent stales but 
 usually failed to parse diagrams with low visual quality. Moreover, there are not any diagram-related annotations in some datasets (e.g., GeoQA \cite{GeoQA}) to train the parsing model. Therefore, the parsing methods are ineffective to find symbolic characters in our work. On the other hand, we may adopt an optical character recognition (OCR) tool MathPix \footnote{https://mathpix.com/} to directly recognize symbolic characters in the diagram image~\cite{Inter-GPS}. However, such OCR tools usually output poor recognition results on low-quality images. To detect symbolic characters in the low-quality diagram images, we utilize the powerful Segment Anything Model (SAM)~\cite{SAM} in our work.

\subsection{Multimodal Reasoning}
 In visual question answering (VQA), multimodal reasoning is significant to provide joint information over different modal features \cite{RWmultimodal0}. 
 Some previous works on VQA jointly encode multimodal information through an implicit reasoning framework \cite{RWmultimodal1}.
 Furthermore, \cite{RWmultimodal3} converts visual features into specific languages and utilizes symbolic reasoning to answer the question.   
 Similar to VQA, solving geometry problems also requires multimodal reasoning on features from both text description and diagram image.
 However, different from general VQA tasks like objects counting \cite{RWmultimodalCount} and natural language-based answering \cite{RWmultimodalVisual}, solving geometry problems usually further relies on external math theorems and deductive calculation \cite{GeoQA}, therefore, it is difficult to directly adopt general VQA multimodal models.

\subsection{Self-Supervised Learning}
 Transformer with self-supervised learning has achieved great success in natural language processing (NLP), such as masked language modeling training in BERT \cite{bert}. 
 Motivated by the success of Transformer-based models in NLP, \cite{vit} proposed a Vision Transformer model with self-supervised learning by masked patch prediction. ViT  is usually not easy to optimize, which requires a huge number of training resources and high computation costs. 
 To effectively train ViT, \cite{vitmae} proposed the ViTMAE method which randomly masked a number of patches in the pixel space, then learned to reconstruct these patches by an autoencoder method. 
 Inspired by ViTMAE, we utilize a masked image modeling task to pre-train our Transformer-based diagram encoder, where we specifically design the mask percentage as our diagram images mainly contain straight lines and annotated characters without those sufficient textures in natural images.

\section{Methodology}
In this paper, we develop a novel Symbolic Character-Aware Geometry Problem Solver, abbreviated as SCA-GPS. Figure~\ref{fig:overview} shows one overview of our proposed SCA-GPS framework.
Given a geometry problem $P = [P_T, P_D]$, we first adopt two encoders to extract features from text $P_T$ and diagram $P_D$ respectively. To be noted, exact numbers in the text description are meaningless during the model reasoning, which is only useful for the final answer calculation. Therefore, we replace each number in the text $P_T$ with an independent variable such as $N_1, N_2$. Next, we fuse both features from the text and diagram in a multi-modal reasoning block to output the final feature representation, which is then input to the decoder to predict the program codes. Finally, these codes are executed to output the final answer, where the variables will be replaced back by the corresponding numbers. To better optimize the diagram encoder, we propose two auxiliary tasks to train it in advance. In the text encoder, we merge symbolic characters to form a holistic semantic unit based on the aligned features with the diagram image. In the following, we will describe each block of our SCA-GPS framework in details.

\subsection{Diagram Encoder}
\label{Sec_diagram}
We engage a ViT \cite{vit} model to encode the geometry diagram into a sequence of features. 
Given a geometry diagram image $P_D$, we first divide it into regular non-overlapping $n \times n$ patches, then we input these patched images to the ViT model individually and take the output of the last hidden layer in the ViT encoder as the final diagram encoded features $F_{D} = [f_{D}^1,f_{D}^2,...,f_{D}^{m}], m=n\times n$.
To train the ViT model, we design two auxiliary tasks including masked image modeling and multi-label classification. Once this optimization is done, we do not update the diagram encoder during the training of the whole GP solver, saving lots of training resources. To be noted, no human annotation is needed for both auxiliary tasks, which makes it convenient to adopt. The details will be described in the following.

\subsubsection{Masked Image Modeling.}
  Motivated by ViTMAE, we adopt a masked image modeling (MIM) task to help the diagram encoder effectively extract geometry features and learn geometry primitives.
  In the original configuration of ViTMAE, an image is split into $16 \times 16$ patches and the mask ratio is set as 0.75. 
  However, we think this configuration is not suitable for processing geometry diagrams, where most pixels in the diagram image are purely white and have less visual information than the natural images.
  Meanwhile, geometry primitives (e.g., points, lines, arcs, and circles) have stronger constraints. Thus, we intend to make ViT extract accurate geometry features. As such, we set the split strategy as $8 \times 8$ (64 patches for each image) and set a much smaller mask ratio of 0.2 in our experiments.
  Similar to ViTMAE \cite{vitmae}, we calculate the mean squared error (MSE) between the reconstructed and original image patches at pixel-level as the loss function $\mathcal{L}_{mim}$ for the MIM task, which can be formulated by
        \begin{equation}
                    \mathcal{L}_{mim} =\frac{\sum_{i=1}^N  L_i \cdot M_{i}}{\sum_{i=1}^N  M_{i}}, \textbf{where } 
        L_i=\frac{1}{C} \sum_{j=1}^C \left({P}_{i j}-T_{i j}\right)^2.
        \end{equation}
In the above, ${P}_{i j}$ is the prediction of the $j$th pixel value in patch $i$ while $T_{i j}$ is the target pixel,  $C$ is the number of pixels in each patch, $L_{i}$ is the loss of patch $i$, and $M_{i}$ is the mask of each patch to only calculate the loss on removed patches.

\begin{figure}
    \centering
    \includegraphics[scale=0.9]{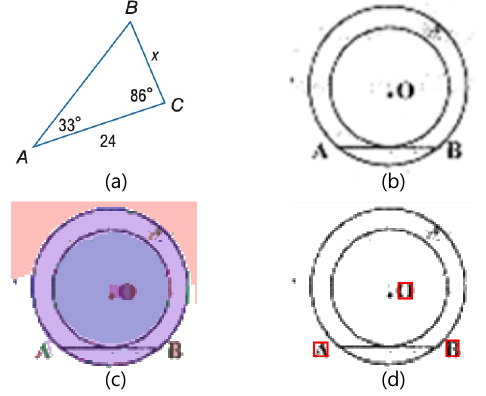}
    \caption{Examples of geometry diagrams with different image qualities. (a) high-quality image from Geometry3K~\cite{Inter-GPS}, (b) low-quality image from GeoQA~\cite{GeoQA}, (c) the segmentation mask of (b) by SAM~\cite{SAM}, (d) our symbolic characters detection results on (c). }
    \label{fig:SAM-VIT}
\end{figure}

\subsubsection{Multi-label Classification.}
 Symbolic characters play an important role in bridging text description with diagram images. Previous works~\cite{Inter-GPS, PGDP} usually parse the diagram into textual formal languages to connect the text description. However, these methods only perform well for high-quality diagram images with consistent stales (e.g., Figure~\ref{fig:SAM-VIT} (a)) but usually fail to parse diagrams with low visual quality (e.g., Figure~\ref{fig:SAM-VIT} (b)).  To detect symbolic characters in low-quality diagram images, we utilize the powerful SAM~\cite{SAM} in our work. In detail, we first input the diagram image into SAM to obtain the segmentation masks (Figure \ref{fig:SAM-VIT} (c)), consisting of numerous objects named 'cut-outs'. Next, we manually filter those objects by checking both the aspect ratio of its bounding box and the size ratio to the original diagram size. For those adjacent boxes for each character, we evaluate the Intersection of Union (IoU) of objects to combine overlapped bounding boxes and finally output the symbolic characters detection results (Figure \ref{fig:SAM-VIT} (d)).  

 Based on the detection results and the original diagram encoded features $F_{D} = [f_{D}^1,f_{D}^2,...,f_{D}^{m}]$, we can obtain a masked diagram features $\tilde{F}_{D} = [\tilde{f}_{D}^1,\tilde{f}_{D}^2,...,\tilde{f}_{D}^{m}]$. Specifically, $\tilde{f}_{D}^i=f_{D}^i$ if this patch contains a detected symbolic character, otherwise $\tilde{f}_{D}^i=\textbf{0}$. Then, we input the masked feature $\tilde{F}_{D}$ into an attention reduction model~\cite{mcan}:
 \begin{equation}
    \label{equ_att_reduction}
    \alpha=\operatorname{softmax}\left(\operatorname{MLP}\left({\tilde{F}_{D}}\right)\right), {f^{*}}=\sum_{i=1}^m \alpha_i {\tilde{f}_{D}^{i}},
 \end{equation}
 to obtain attended feature $f^{*}$ for the following multi-label classification task.

In the classification, we input $f^{*}$ to a linear layer with the Sigmoid activation function to give the prediction score of symbolic characters. 
As the dataset does not provide symbolic characters annotation for each diagram image, we rely on a weakly supervision method for our multi-label classification. In details, we define all symbolic characters in the corresponding text description are positive classes while the others are deemed negative. 
The training loss $\mathcal{L}_{char}$ is defined by the binary cross entropy: 
  \begin{equation}
    \mathcal{L}_{char}=\frac{1}{N} \sum_{n=1}^N-w_n\left[y_n\cdot \log x_n+(1-y_n) \cdot\log (1 - x_n) \right],
  \end{equation}
where $N$ is the size of the full set (i.e. 27 in our experiments including 26 alphabets a-z and one additional class for non-character objects), $x_n$ and $y_n$ represent the prediction score and label for the $n$-th sample, respectively. It is noted that most of the geometry problems mainly consist of the first several characters (e.g. A, B , C, D) while others rarely exist (e.g. T, U, V, W). Therefore, we add a weight $w_n$ to balance these classes. 
It is possible that there are some noisy samples if the characters in the text do not have any geometry meaning with the diagram image. 
However, this method does not require human annotation workload and also shows effectiveness in the experiments.  
Finally, we optimize our diagram decoder by the combination of both $\mathcal{L}_{mim}$ and $\mathcal{L}_{char}$ as follows:
    \begin{equation}
    \mathcal{L}_{Aux} = \mathcal{L}_{mim} +  0.1   \mathcal{L}_{char} .
    \end{equation}

\subsection{Text Encoder}
 To comprehensively understand symbolic characters with corresponding diagram features, we propose a character-aware text encoder to better encode text descriptions.
 First, we tokenize text $P_T$ into a token sequence $T_{token}=[t_1,t_2,...,t_l]$ where $l$ is the length of text, then an embedding layer is used to obtain initial embedded features $E_T=Embedding(T_{token})$. To highlight the symbolic characters in the text description, we define a mask $M_{align}$ in advance, where $M_{align}^i=1$ if the $i$th character is a symbolic character, otherwise $M_{align}^i=0$.
 In addition, we propose to merge those consecutive symbolic characters (equal or longer than two consecutive tokens), which are defined by another mask $M_{merge}$. Specifically, $M_{merge}^i=1$ and $M_{merge}^{i+1}=1$ if $M_{align}^i=1$ and $M_{align}^{i+1}=1$, otherwise $M_{merge}^i=0$. 
 
   \begin{figure}
     \centering
     \includegraphics[scale=0.9]{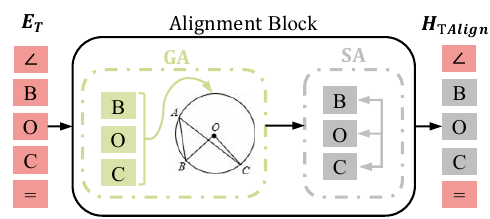}
     \caption{Flowchart of the character-diagram alignment block. Each block consists of a Guided-Attention (GA) and a Self-Attention (SA) layer, the module is stacked by 2 blocks.}
     \label{fig:charattention}
  \end{figure}

\subsubsection{Character-Diagram Feature Alignment}
To align symbolic character features with geometry diagrams, we come up with an attention-based character-diagram alignment module. Motivated by \cite{mcan}, our alignment module consists of two Self-Attention (SA) and Guided-Attention (GA) blocks as shown in Figure \ref{fig:charattention}.
In general, the alignment process is formulated by:
$$H_{TAlign}=Align(\tilde{E}_T, F_D),$$ where $\tilde{E}_T$ and $F_D$ denote the masked embedded text feature and encoded geometry diagram feature, respectively. 
The masked text features $\tilde{E}_T$ are obtained by masking the original embedding feature ${E}_T$ according to the symbolic character mask $M_{align}$. Specifically, $\tilde{E}_{T}^i=E_{T}^i$, if $M_{align}^i=1$, otherwise $\tilde{E}_{T}^i=\textbf{0}$.  
In details, the alignment process can be defined as:
    \begin{equation}
            Align(\tilde{E}_T, F_D) = Block_2(Block_1(\tilde{E}_T, F_D), F_D),
    \end{equation}
where each block consists of one GA followed by a SA:
\begin{equation}
        Block_n = SA(GA(\tilde{E}_T, F_D)).
    \end{equation}
In each alignment block, text features first apply Guided-Attention to $F_D$, to acquire geometry visual features into symbolic characters:
    \begin{equation}
    A_{GA}=GA(Q_T, K_D, V_D),
    \end{equation}
where query from text (i.e., $Q_T=\tilde{E}_T$), key and value from the diagram (i.e., $K_D=F_D$, and $V_D=F_D$). Next, we use Self-Attention to share relative information between symbolic characters
    \begin{equation}
    A_{SA}=SA(Q_T, K_T, V_T),
    \end{equation}
where the query, key, and value take the output of the GA layer (i.e., $A_{GA}$). For both GA and SA, we use a scaled dot-product attention mechanism: 
    \begin{equation}
        A(Q, K, V)=\operatorname{softmax}\left(\frac{Q K}{\sqrt{d}}\right) V,
    \end{equation}
where $d$ is the dimension of features. 

To be noted, only the features of those masked symbolic characters are updated in our Character-Diagram Feature Alignment module, for example, the features of characters 'B', 'O', and 'C' in Figure~\ref{fig:charattention}. 
Once we complete the update of symbolic character features, the other character features will be replaced by the original features (i.e., $H_{TAlign}^i={E}_T^i$ if $M_{align}^i=0$).

\begin{figure}
      \centering
      \includegraphics[scale=0.9]{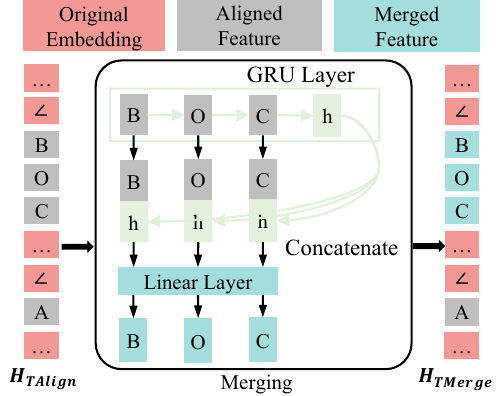}
      \caption{Flowchart of our symbolic character merging module. Each subsequence of consecutive symbolic characters is fed into a GRU layer, resulting in a final hidden state. This hidden state is then concatenated with the corresponding aligned features to generate a merged feature representation.}
      \label{fig:txtencoder}
  \end{figure} 

  \subsubsection{Symbolic Characters Merging}
  Inspired by~\cite{wu2021math} that explicitly processes numerical values in MWP, we propose a symbolic character merging module. After aligning symbolic characters with diagram features, we apply a merging operation to form the united semantic information of individual characters which are then fused with each individual original feature.

Figure~\ref{fig:txtencoder} shows a general flowchart of symbolic character merging. 
Based on the merging mask $M_{merge}$, we first extract consecutive symbolic characters (i.e., a consecutive sub-sequence $M_{merge}^i=1$) features from the aligned text feature $H_{TAlign}$ to form a sub-sequence features $H_{subseq}$. 
For example, only the features of characters 'B', 'O', 'C' in Figure~\ref{fig:txtencoder} are extracted to construct $H_{subseq}$.  
Next, we input $H_{subseq}$ to a GRU layer, and take the last hidden state of GRU as the merged feature representation $h$: $$h = GRU(H_{subseq}).$$
Then, we concatenate $h$ to each original token feature in $H_{subseq}$ and input to a linear layer to fuse features of both merged characters and original individual characters, which can be formulated by 
$$I_{subseq}=ReLU(Linear(i_0\oplus_{}{}h, i_1\oplus_{}{}h,..., i_{n-1}\oplus_{}{}h)),$$ where $i_n$ is the individual symbolic character feature in $H_{subseq}$.
Next, we use $I_{subseq}$ to replace the corresponding character features in $H_{TAlign}$, while the remained character features in $H_{TAlign}$ are not updated, forming our final merged text feature $H_{TMerge}$.
The detailed symbolic character merging algorithm can be found in  Algorithm \ref{alg:algorithm}, where $\oplus_{}{}$ denotes concatenation.

\begin{algorithm}[]
    \caption{Symbolic Character Merging}
    \label{alg:algorithm}
    
    \begin{algorithmic}[1] 
        \STATE \textbf{Input}: $H_{TAlign}, M_{merge}$ 
        \STATE \textbf{Output}: $H_{TMerge}$ 
        \STATE $H_{TMerge} = H_{TAlign}$ 
        \STATE Let $Count=0$.
        \FOR{$i$ in range(length($M_{merge}$))}
        \IF {$m_i$ == 1}
        \STATE $Count+=1$
        \IF {$m_{i+1}$ == 0} 
            \STATE $H_{subseq}=H_{TAlign}[i-Count+1:i+1]$
            \STATE $h = GRU(H_{subseq})$ 
            \STATE $h_{cat}=[H_1\oplus_{}{}h,...,H_n\oplus_{}{}h],n=Count$
            \STATE $h_{fuse} = Linear(h_{cat})$
            \STATE $H_{TMerge}[i-Count+1:i+1] = h_{fuse}$
            \STATE $Count$=0
        \ENDIF
        \ENDIF
        \ENDFOR
        \STATE \textbf{return} $H_{TMerge}$
    \end{algorithmic}
\end{algorithm}

  \subsubsection{Dual Text Encoder}
  To better obtain text features, we adopt a dual text encoder by following the work~\cite{GeoQA+}. Concretely, we utilize an LSTM as one baseline text encoder in our proposed symbolic character-aware model to obtain the text features $H_{TMerge}$. 
  As shown in Figure~\ref{fig:overview}, we adopt a RoBERTa~\cite{roberta} as an additional text encoder, to learn semantic information of complicated problem descriptions, and the output is represented by $H_{bert}$. To be noted, we do not integrate our proposed symbolic character-aware method in  RoBERTa. 
  Finally, we combine the features of $H_{bert}$ and $H_{TMerge}$ together by a linear fusion layer to get the final text features $F_T=[H_{TMerge} + H_{bert}]$. The text feature $F_T$ together with the encoded diagram feature $F_D$ will be fed to the following multi-modal reasoning module.

\subsection{Multimodal Reasoning}
 Solving a GP usually requires jointly understanding multi-modal features of both the problem text and diagram. To this end, we adopt a commonly used multi-modal co-attention module \cite{mcan} to align $F_T$ and $F_D$. 
 The co-attention module has 12 Self-Attention (SA) units and 6 Guided-Attention (GA) units to fuse text and diagram features.
 Text feature $F_T$ first goes through 6 SA units to model internal information and output the hidden state $H_T$ from the last SA unit.
 Next, $H_T$ and $F_D$ are fed into a text-image guided attention stack (6 SA-SGA units) as guiding features to fuse multi-modal information, the output $H_D$ represents cross-modal features.
 Considering text description has reliable information for GP understanding \cite{GeoQA}, we concatenate the text feature $F_T$ and the cross-modal feature $H_D$ to form $F_R=[F_T, H_D]$ as the final output of the multimodal reasoning module, which will be fed to the following decoder in sequence.

 Meanwhile, we aim to obtain an overall representation feature vector $f_R$ for the initial state of the following sequence decoder. To this end, we first concatenate the first and the last token features in $F_T$, and feed them into a linear layer to get a vector $f_T$ for overall text features. Similarly, we map the feature sequence $H_D$ into an overall representation $ h_D$ via an attentional reduction network~\cite{mcan} shown in Equation (\ref{equ_att_reduction}). 
 Then, we concatenate $f_T$ with $h_D$ to finally obtain the overall multi-modal feature vector $f_R=[f_T, h_D]$ by a linear transformation.

\subsection{Decoder and Executor}
 To acquire the executing program code, we use an LSTM~\cite{LSTM} based decoder with attention mechanism \cite{attention} to sequentially generate target program code based on the multi-modal reasoning output $F_R$. 
 Given a target program sequence ${y_t}(1 \leq t \leq T)$ as the ground truth, we have $P_t$ as the next program token to be predicted and $s_t$ as the LSTM hidden state at time step $t$. 
 At the first step $0$, we feed $f_R$ into a linear layer to get the initial hidden state $s_0$. Subsequently, each $s_t$ attends to $F_R$ and concatenates the attended result to predict token $P_t$ with the softmax function.

 The entire model except for the diagram encoder\footnote{The diagram encoder is pre-trained by its two auxiliary tasks in advance as shown in section~\ref{Sec_diagram}.} is optimized by negative log-likelihood loss function:
 \begin{equation}
 \mathcal{L}_g(\boldsymbol{\theta})=\frac{1}{T} \sum_{t=1}^T \log P_t\left(y_t \mid \boldsymbol{x}, y_1, \ldots, y_{t-1} ; \boldsymbol{\theta}\right),     
 \end{equation}
 where $\boldsymbol{\theta}$ represents the parameters of model, $\boldsymbol{x}$ represents text and diagram features. During the testing, we apply a beam search strategy with beam size $B$ during the decoding process, and output $B$ program sequences finally.

 Once we get the generated program sequence, we can run a program executor on the sequence by handcraft math calculate functions to give the final numerical answer. To be noted, the program executor is dataset-oriented. Different geometry problem datasets are normally associated with different executors \cite{GeoQA,Inter-GPS}.
 In the dataset GeoQA, the executor chooses the answer from the given choices once the output sequences can get the same value; otherwise, it will mark the problem as "No Result".
 In the dataset Inter-GPS, the rule-based geometry solver follows the program sequence to apply a series of handcraft theorems over parsed formal languages. If the program cannot get the problem goal, it will use the "Low-first Search Strategy" to solve the problem. Therefore, there will not output "No Result" for Inter-GPS.

\section{Experiments}

 \subsection{Datasets and Implementation Details}
 To verify the effectiveness of our method, we conduct experiments on two benchmark datasets: GeoQA \cite{GeoQA} and Geometry3K \cite{Inter-GPS}.
 On one hand, GeoQA consists of 3,499 training samples, 745 validation and 754 test samples. These geometry problems (GP) are divided into three types: Angle, Length, and Other. In the GeoQA, there are diverse GP styles and low-quality diagram images. On the other hand, Geometry3K contains 3,002 problems in total, but there are only 2,103 problems with solving program annotations, including 1,311 training samples, 191 validation and 601 test samples. In addition, there is another public GP dataset~\cite{introEdu}, however, it only has 68 training samples, which is too small to be available for evaluation. 
 During training, the learning rate of RoBERTa is set to $2e^{-5}$, and that of multi-modal reasoning and character-diagram alignment module is set to $e^{-5}$, while $e^{-3}$  is used for the other modules. In ViT diagram encoder, we set the patch size as 28, and the total patch number as ($8\times 8$). The learning rate for training ViT encoder is $e^{-4}$. The mask ratio of ViT is 0.2 when we conduct the MIM task. 
 All experiments were conducted on an NVIDIA Quadro RTX8000 GPU with the batch size 32. Adam is chosen as the optimizer, the beam size $B=10$, and the input image is resized to $224\times 224$.
 The training epoch of the whole solver is 100, while it is 300 for ViT in the auxiliary tasks.

\begin{table}
    \centering
    \caption{Solving accuracy (\%) of various methods on the GeoQA dataset. The first two rows indicate the performance solved by humans.}
    \begin{tabular}{l|cccc}
    \hline
    Method           & Total     & Angle     & Length    & Other    \\
    \hline
    Text-Only        & 63.0      & 58.0      & 71.7    & 55.6       \\
    Text-Diagram     & 92.3      & 94.2      & 90.5      & 87.0     \\
    \hline
    Seq2Prog         & 54.2      &  66.4   &  38.5      & 42.6           \\
    BERT2Prog        & 53.7      &  65.5   &  39.6      & 37.0            \\
    RoBERTa2Prog     & 54.4      &  65.2   &  40.2      & 44.4            \\
    NGS \cite{GeoQA}       & 60.0     & 71.5      & 48.8     & 29.6     \\
    DPE-NGS \cite{GeoQA+}  & 62.7     & 74.9      & 47.7     & 50.0    \\
    SCA-GPS (Ours)          &\textbf{64.1}  &\textbf{74.9}      &\textbf{50.1}  &\textbf{55.5} \\
    \hline
    \end{tabular}
    \label{tab:geoqaResult}
\end{table}

\subsection{Experimental Results}
In this section, we will show the experimental results on GeoQA and Geometry3K, respectively.

\subsubsection{GeoQA Dataset}
We compare our proposed SCA-GPS model with the previous works on GeoQA dataset as shown in Table~\ref{tab:geoqaResult}. To be noted, some previous works are not compared in our experiments due to requiring additional inputs~\cite{javaGeo} or no open-source codes~\cite{sachan2017learning}. In addition, we also compare three general sequence-to-sequence models without any special GP solving techniques, including an LSTM-based model (Seq2Prog), a BERT-based model (BERT2Prog) \cite{bert} and a RoBERTa-based model (RoBERTa2Prog) \cite{roberta}. From Table \ref{tab:geoqaResult}, we can see that the general sequence-to-sequence models can solve GPs, but the accuracy is limited. All of the GP specific methods improve the solving accuracy obviously, and our method achieves the best accuracy of 64.1\%, substantially better than the second-high accuracy 62.7\%~\cite{GeoQA+}, even higher than the human solving with Text-Only. By comparing the accuracy on different GP types, our method also performs the best. All of these demonstrate the effectiveness of our proposed SCA-GPS model. 
However, our method is still far behind the human solving with Text-Diagram, indicating that this task is still far from being solved. 

In addition, we compare the performance of the "No Result" metric with the other methods in Table \ref{tab:geoqaNoresult}. We can see that our model can achieve the smallest number of "No Result" answers, indicating that our model learns more reasonable solving logic, then most of the predicted programs can obtain an answer provided in the question even not exactly correct. Furthermore, a lower "No Result" score means our model has learnt about how to deal with different problems, not just simply using general previously learned solutions for similar problems.

\begin{table}
    \centering
    \caption{Percentage of 'No Result' problems predicted by different models on the GeoQA dataset.}
    \begin{tabular}{lcc}
        \hline
        Model                   & Total (\%)     & No Result (\%) \\
        \hline
        NGS \cite{GeoQA}        & 60.0          & 14.86    \\
        DPE-NGS \cite{GeoQA+}   & 62.7          & 12.68    \\
        {SCA-GPS(Ours)}         &\textbf{64.1}  & \textbf{11.94}    \\
        \hline
    \end{tabular}
    \label{tab:geoqaNoresult}
\end{table}

\begin{table}
    \centering
    \caption{Experiments on Geometry3K. * denotes our reproduced results based on the open-source code.}
    \begin{tabular}{lcc}
        \hline
        Dataset  & Accuracy (\%) & Steps (Avg) \\
        \hline
        NGS \cite{GeoQA}            & 76.2              & 6.2      \\
        Inter-GPS* \cite{Inter-GPS} & 76.3              & 7.1       \\
        SCA-GPS (Ours)               & \textbf{76.7}     & \textbf{6.0}      \\
        \hline
    \end{tabular}
    \label{tab:geo3k}
\end{table}

\subsubsection{Geometry3K}
We also conduct experiments on Geometry3K to predict the theorem solving sequence. 
As our method can not parse the Geometry3K dataset into formal languages, we leverage the ground truth of parsed results provided by the dataset as problem text input, and focus on comparing the theorem predict task performance with Inter-GPS~\cite{Inter-GPS}. 
As general sequence-to-sequence models commonly perform poor for solving GP as shown in Table~\ref{tab:geoqaResult}, we do not compare them in this experiment. 
As shown in Table~\ref{tab:geo3k}, our method can achieve marginally higher accuracy than the Inter-GPS method, but with fewer solving steps. 
Different from the theorem predictor in Inter-GPS, our method and NGS~\cite{GeoQA} encode diagram visual features, which is helpful for reducing solving steps. 
Our approach also explicitly processes symbolic characters through alignment and merging, thereby enhancing the model's comprehension of problem descriptions.

\subsection{Ablation Study}

\subsubsection{Auxiliary Tasks of Diagram Encoder} To verify the effectiveness of our proposed auxiliary tasks for the training diagram encoder, we conduct a series of experiments: 
 (1) training the encoder by the masked image modeling (MIM) task from a pre-trained model with the vanilla ViTMAE configuration (i.e., $16\times16$) on ImageNet;
 (2) training the encoder by the masked image modeling (MIM) task from random initialization with $8\times8$ patching;
 (3) training the encoder by the multi-label classification (MLC) task from random initialization;
 (4) training the encoder by a combination of MIM and MLC tasks.
 From Table \ref{tab:ablation}, we can see that directly fine-tuning the vanilla ViTMAE pre-trained model with $16\times 16$ patch numbers has lower accuracy. This indicates geometry diagrams are significantly different from general pre-training images, which require training a geometry diagram-specific ViT model by auxiliary tasks.
 Furthermore, experiment (2) shows that MIM on an $8\times 8$ patch strategy is more suitable for our SCA-GPS, while the multi-label classification (MLC) task seems insufficient for diagram understanding as shown in the experiment (3). However, by adding the MLC with the MIM task as shown in the experiment (4), the accuracy is increased from 63.3\% to 64.1\%, which indicates that both tasks are complementary and effectively enhance the diagram understanding ability.

\begin{table}
    \centering
    \caption{Accuracy of various diagram encoder auxiliary tasks in our SCA-GPS method on the GeoQA dataset.}
    \begin{tabular}{lcc}
        \hline
        Setting  & Accuracy(\%) \\
        \hline
        (1) Vanilla Pre-trained ViTMAE      &  61.6            \\
        (2) $8\times 8$ ViTMAE with MIM     &  63.3           \\
        (3) $8\times 8$ ViTMAE with MLC     &   60.4          \\
        (4) Jointly train MIM and MLC       &  \textbf{64.1}     \\
        \hline
    \end{tabular}
    \label{tab:ablation}
\end{table}

\begin{table}
    \centering
    \caption{Accuracy of various symbolic character processing techniques in our SCA-GPS method on the GeoQA dataset.}
    \begin{tabular}{lcc}
        \hline
        Setting                         &   Accuracy(\%) \\
        \hline
        (5) w/o Align and Merge         &   61.7     \\
        (6) With Alignment              &   62.5     \\
        (7) With Merging                &   63.7       \\
        (8) With Alignment and Merging  &  \textbf{64.1}    \\
        \hline
    \end{tabular}
    \label{tab:ablation-txt}
\end{table}

\subsubsection{Symbolic Character-Aware Modules.}
To verify the effectiveness of our symbolic character-aware modules, we conduct the following experiments:
(5) A baseline text encoder simply consists of RoBERTa and LSTM;
(6) Adding alignment in the text encoder;
(7) Adding merging in the text encoder;
(8) Adding both symbolic character alignment and merging.
As depicted in Table~\ref{tab:ablation-txt}, the baseline text encoder achieves lower accuracy. 
Moreover, the alignment modules increase performance by connecting visual information from diagrams to their corresponding textual components. 
According to experiment (7), our merging strategy effectively enhances solving accuracy through a holistic comprehension of problem textual descriptions. 
By incorporating both symbolic character alignment and merging mechanisms (setting (8)), we can attain optimal performance, demonstrating that the synergistic interaction between these two mechanisms is effective in tackling complex problems.

\subsection{Case Analysis}
\begin{figure}
    \centering
    \includegraphics[width=\linewidth]{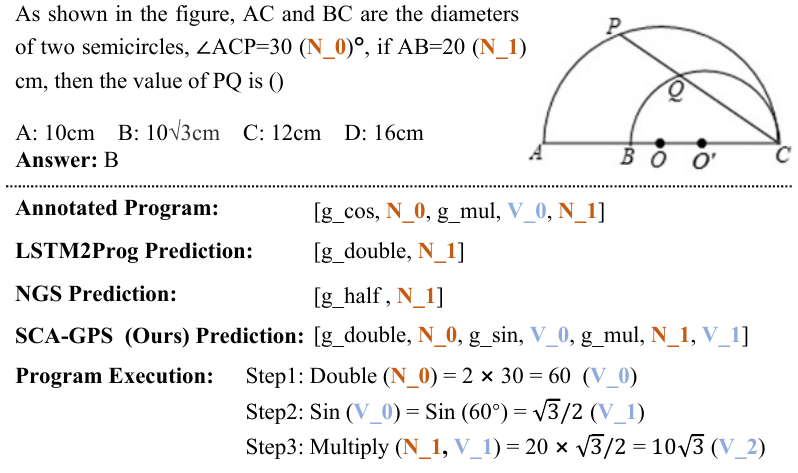}
    \caption{ One illustrative example of comparing predictions from different solving methods.}
    \label{fig:case}
\end{figure}

As shown in Figure \ref{fig:case}, we select a typical geometry problem sample for analysis. 
This problem requires geometry understanding ability and uses trigonometric functions to get a ratio for answering questions.
The LSTM-based method and NGS model give an incorrect answer with a simple but wrong program sequence. 
This is because they lack generalization ability and prefer to solve a problem with simple sequences, while our SCA-GPS model generates a sequence that can obtain the correct answer. 
Despite our generated sequence is not exactly the same as the annotated program, we can also obtain a correct answer, indicating that our method learns generalization ability from the training data. 
This observation may be attributed to the fact that the model infers the degree of this angle from similar problems during the training process. 
As a result, $\angle ACP=30^{\circ}$ is inferred, next $\angle PAC=60^{\circ}$ is obtained, then the executor uses the sine function to multiply with the length of $AB$ to finally obtain $PQ$. 
This also indicates that our model learns the symbolic character features in diagrams without requiring additional annotations.

\section{Conclusion}
 In this paper, we propose a symbolic character-aware model (SCA-GPS) for solving geometry problems (GP). This is a challenging task as it usually involves multimodal information from both text description and diagram image in a GP. 
 Our proposed SCA-GPS model is based on a general encoder-decoder framework, where we adopt two separate encoders for problem text and diagram image respectively. 
 In the diagram encoder, we introduce two tasks for pre-training the ViT-based model. 
 The first task leverages the masked image modeling target to better learn the features within geometric diagrams, while the second task is weakly supervised multi-label classification to predict symbolic characters in the diagram.  
 With respect to the text encoder, we emphasize the significance of symbolic characters and propose to merge them with aligning features from the diagram encoder. 
 Subsequently, the encoded features from both problem text and diagram are fused through a multi-modal reasoning framework, which is then input to a decoder to generate the final program code sequence. 
 Extensive experiments on two benchmark datasets, GeoQA and Geometry3K, demonstrate the effectiveness of the proposed model. 
 In the future, we aim to integrate external geometric theorems into our model to further improve its accuracy and enhance interpretability.

\begin{acks}
This research was funded by National Natural Science Foundation of China (NSFC) no.62276258, Jiangsu Science and Technology Programme (Natural Science Foundation of Jiangsu Province) no. BE2020006-4, European Union’s Horizon 2020 research and innovation programme no. 956123, and UK EPSRC under projects [EP/T026995/1]
\end{acks}

\newpage
\bibliographystyle{ACM-Reference-Format}
\balance
\bibliography{sample-base}


\end{document}